\documentclass[10pt,twocolumn,letterpaper]{article}

\usepackage{cvpr}              

\usepackage{graphicx}
\usepackage{amsmath}
\usepackage{amssymb}
\usepackage{booktabs}
\usepackage{bm}
\usepackage[normalem]{ulem}
\usepackage{color}
\usepackage{diagbox}
\usepackage{multirow}
\usepackage{arydshln}
\usepackage{cellspace}
\usepackage[accsupp]{axessibility}
\definecolor{brown}{RGB}{195, 90, 32}

\definecolor{blue}{RGB}{0, 0, 255}

\definecolor{purple}{RGB}{128,0,128}

\usepackage[pagebackref,breaklinks,colorlinks]{hyperref}

\usepackage[capitalize]{cleveref}
\crefname{section}{Sec.}{Secs.}
\Crefname{section}{Section}{Sections}
\Crefname{table}{Table}{Tables}
\crefname{table}{Tab.}{Tabs.}

\def\cvprPaperID{3869} 
\def\confName{CVPR}
\def\confYear{2022}

\begin{document}
\setlength{\aboverulesep}{0pt}
\setlength{\belowrulesep}{-1pt}
\title{Learning to Imagine: Diversify Memory for \\Incremental Learning using Unlabeled Data}

\author{Yu-Ming Tang$^{1,3}$\thanks{Both authors contributed equally}
\qquad Yi-Xing Peng$^{1,3}$\footnotemark[1]
\qquad Wei-Shi Zheng$^{1,2,3}$\thanks{Corresponding author}\\
{$^1$School of Computer Science and Engineering, Sun Yat-sen University, China}\\
{$^2$Peng Cheng Laboratory, Shenzhen, China}\\
{$^3$Key Laboratory of Machine Intelligence and Advanced Computing, Ministry of Education, China}\\
{\tt\small \{tangym9, pengyx23\}@mail2.sysu.edu.cn}
\qquad {\tt\small wszheng@ieee.org}
}
\maketitle

\begin{abstract}

Deep neural network (DNN) suffers from catastrophic forgetting when learning incrementally, which greatly limits its applications. 
Although maintaining a handful of samples (called ``exemplars'') of each task could alleviate forgetting to some extent, existing methods are still limited by the small number of exemplars since these exemplars are too few to carry enough task-specific knowledge, and therefore the forgetting remains.
To overcome this problem, we propose to ``imagine'' diverse counterparts of given exemplars referring to the abundant semantic-irrelevant information from unlabeled data.
Specifically, we develop a learnable feature generator to diversify exemplars by adaptively generating diverse counterparts of exemplars based on semantic information from exemplars and semantically-irrelevant information from unlabeled data. We introduce semantic contrastive learning to enforce the generated samples to be semantic consistent with exemplars and perform semantic-decoupling contrastive learning to encourage diversity of generated samples.
The diverse generated samples could effectively prevent DNN from forgetting when learning new tasks.
Our method does not bring any extra inference cost and outperforms state-of-the-art methods on two benchmarks CIFAR-100 and ImageNet-Subset by a clear margin.

\end{abstract}

\section{Introduction}
\label{sec:intro}

 \begin{figure}
        \centering
        \begin{subfigure}{\linewidth}
        \centering
            \includegraphics[width=\linewidth]{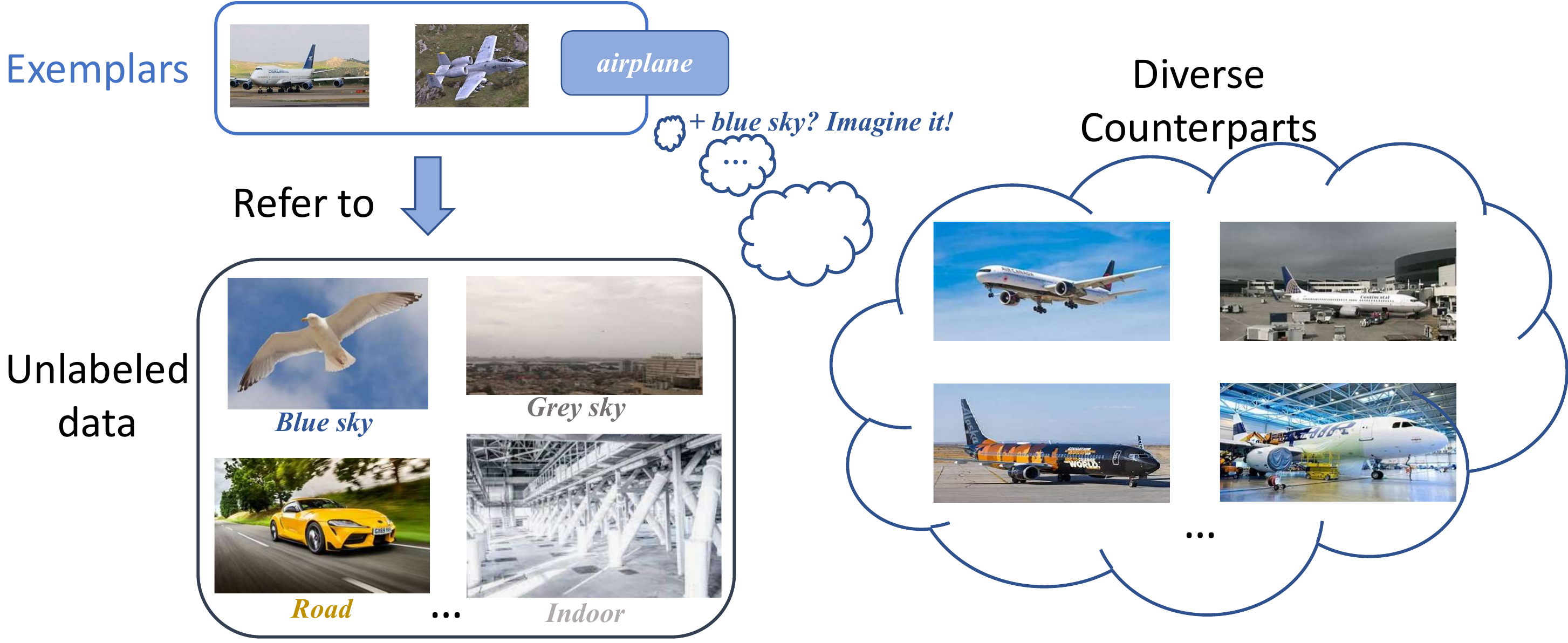}
            \caption{Intuitive understanding of our method.}
            \label{fig:intro_2}
        \end{subfigure}%
        \\ 
        \begin{subfigure}{\linewidth}
        \centering
            \includegraphics[width=\linewidth]{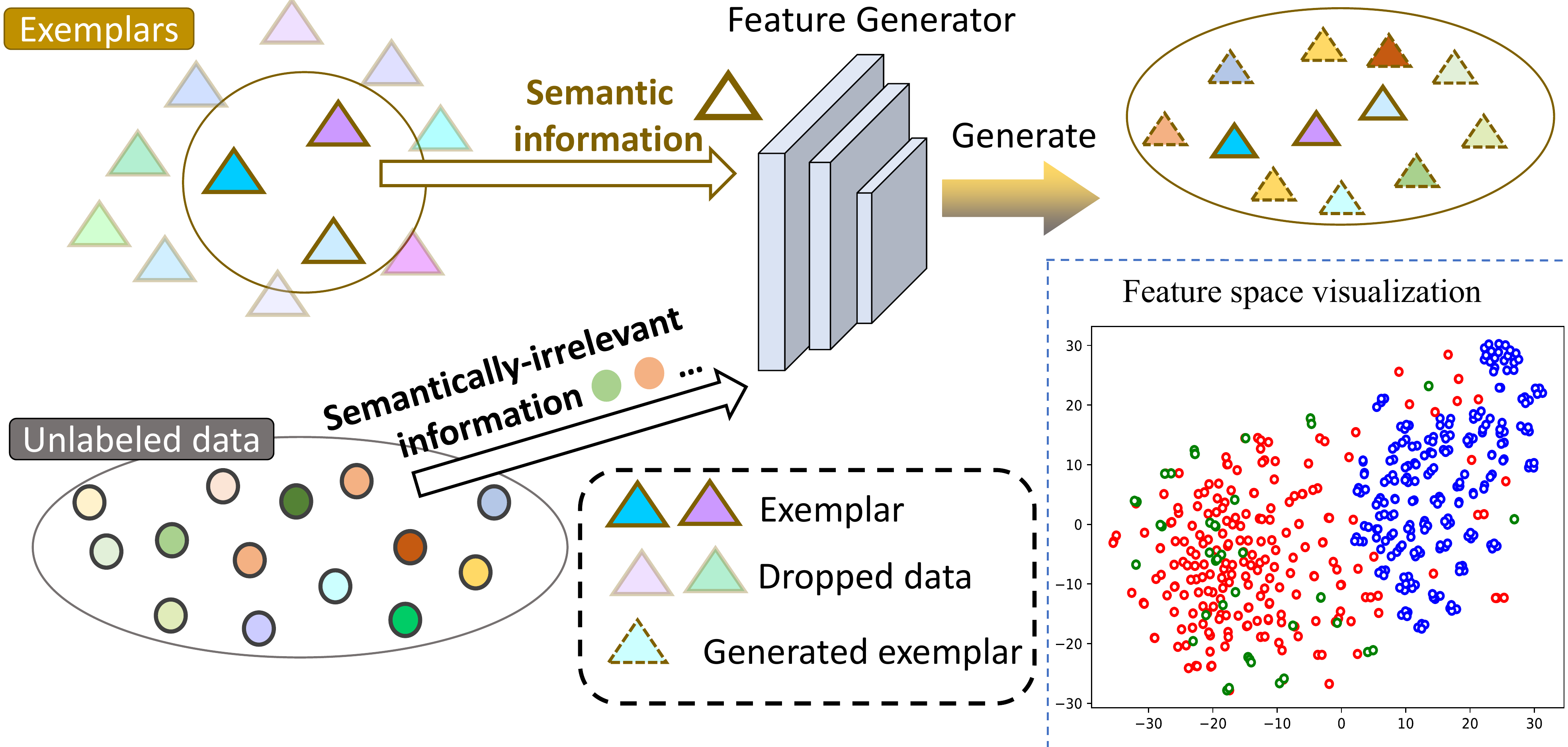}
            \caption{Illustration of our method and TSNE visualization.}
            \label{fig:intro_1}
        \end{subfigure}
        \caption{(a). Our motivation. We propose to `imagine' diverse counterparts of the limited exemplars referring to the semantically-irrelevant information from unlabeled data. 
        (b). We propose a feature generator to `imagine' diverse counterparts
        by adaptively mixing semantic information from exemplars with semantically-irrelevant information from unlabeled data. In the TSNE visualization, blue/green/red dots are the features of one class's generated samples/exemplars/real data. We observe that the exemplars and generated samples cover real data. Best viewed in color. }
        \label{fig:intro}
        \end{figure}
Recent years have witnessed the rapid development of deep neural networks (DNNs) in various tasks\cite{resnet,vgg,alexnet}.
However, when a pretrained deep model learns a new task, it tends to forget the knowledge learned from previous tasks in the absence of the corresponding training data \cite{lwf,french1999catastrophic,mccloskey1989catastrophic,goodfellow2013empirical,PARISI201954}.
Such a \emph{catastrophic forgetting} phenomenon greatly limits the real-world application of deep models because it is impractical to maintain the training data of each task due to privacy concerns and so forth \cite{icarl, bic,castro2018end,ucir,rehearsal,random_path,Ordisco}.

To overcome catastrophic forgetting, incremental learning methods are developed.
Previous works widely adopt rehearsal strategy\cite{icarl,podnet,rainbow,bic,inthewild}: storing a limited quantity of samples called \emph{exemplars} from the original training dataset and reusing them against forgetting when the model learning new tasks.
For instance, RM\cite{rainbow} selects hard samples as exemplars according to the classification uncertainty.  GD\cite{inthewild} distills the knowledge from the old network to the new one based on stored exemplars. BiC\cite{bic} optimizes the classification bias referring to a subset of exemplars.

However, only a handful of exemplars conveying limited variations could be stored due to the reasons such as privacy concerns, which hinders the development of existing methods.
When learning a new task with abundant training data, the exemplars are too few and the model capacity tends to be dominated by the training data of the new task. Although one could emphasize the exemplars during learning, the deep model may overfit the exemplars as shown in recent work\cite{rehearsal}, leading to unsatisfactory performance. 

In this work, we propose a plug-and-play learnable feature generator to adaptively diversify the exemplars by exploiting unlabeled data.
When a deep model learns new tasks, it is easy to collect massive unlabeled data in the real world\cite{dmc,inthewild}. Referring to the abundant semantic-irrelevant information within the unlabeled data, we could learn a feature generator to `imagine' various counterparts for the given exemplars and consequently diversify the exemplars for tackling the forgetting problem (See \cref{fig:intro}). 

Our method adopts a two-stage training schedule. Specifically, we sample a handful of exemplars from the current dataset when a task ends. Before dropping the original dataset, we train the feature generator to generate diverse counterparts of the exemplars based on exemplars and massive unlabeled data. We perform semantic contrastive learning between the generated samples and the original dataset so that (i) the generator could learn to keep the generated samples semantically consistent with the exemplars and (ii) the generated samples are encouraged to be as diverse as possible.
To further facilitate the exploration of semantically-irrelevant information within unlabeled data and generate more diverse samples, we further introduce semantic-decoupling contrastive learning between the generated samples and the unlabeled data.
When a new task starts, the feature generator is frozen and used to generate diverse samples to prevent the deep model from forgetting knowledge of previous tasks.
At this time, the feature generator does \textbf{not} require any gradient and serves as a static non-linear mapping function.
Our method does not bring extra inference costs. The feature generator is discarded and only the vanilla deep model is needed during inference.

Our main contributions are as follows. Firstly, we proposed a learnable feature generator to adaptively generate diverse counterparts of limited exemplars by exploiting the semantically irreverent information in a messy unlabeled dataset. With the diverse generated samples, the model could better overcome forgetting. Our method does not bring extra inference cost and is insensitive to unlabeled data.
Secondly, we introduce semantic contrastive learning and semantic-decoupling contrastive learning to ensure the generated samples are diverse and semantically consistent with given exemplars.
Finally, experimental results show that our method is effective and outperforms existing methods by a clear margin with arbitrary unlabeled data.
\section{Related Work}
\label{sec:related}
Existing class incremental learning methods can be roughly divided into two categories: \textbf{data-driven} methods and \textbf{structure-driven} methods on the basis of alleviating the forgetting problem by optimizing the data supply or changing the network structure.

\noindent\textbf{Data-driven methods.} Existing data-driven methods\cite{icarl,podnet,smith2021always,ucir,Mnemonics,rainbow,castro2018end,inthewild,dmc} focus on the data and the relation of new and old data to tackle forgetting.
Many works use \emph{distillation} to maintain representations of kept data. 
Icarl\cite{icarl} and EE2L\cite{castro2018end} pull output logits between old and new models closer via distillation loss. 
PODNet\cite{podnet} further constrains feature representations on different scales of the network based on the old network. 
UCIR\cite{ucir} utilizes normalized features to distill between old and new models instead of raw features.
Some works use different sources of data to assist training.
GD\cite{inthewild} samples unlabeled data in the wild and define a global distillation loss for anti-forgetting learning.
DMC\cite{dmc} uses an extra unlabeled dataset to align the old representations with the new one via a double distillation loss. 
Many rehearsal-based works focus on memory management.
Mnemonics Training\cite{Mnemonics} proposes a bi-level optimization via meta-learning which makes exemplars trainable. 
RM\cite{rainbow} offers a novel memory management strategy that selects hard samples by checking the classification uncertainty after adding noise into samples. 

Most data-driven methods either focus on memory management or exemplar replay strategy albeit considering the existence of unlabeled data.  
A few previous works including GD\cite{inthewild} and DMC\cite{dmc}, use unlabeled data to assist training.
However, their methods\cite{inthewild,dmc} only simply enforce the logits of unlabeled data outputted from the new model to be consistent with that from old models.
Differently,  our proposed feature generator effectively utilizes the abundant variations within unlabeled data to diversify exemplars, which is orthogonal to existing methods. Besides, our feature generator is learnable, making our memory more adaptive than others.

\begin{figure*}
  \centering
    \includegraphics[width=0.98\linewidth]{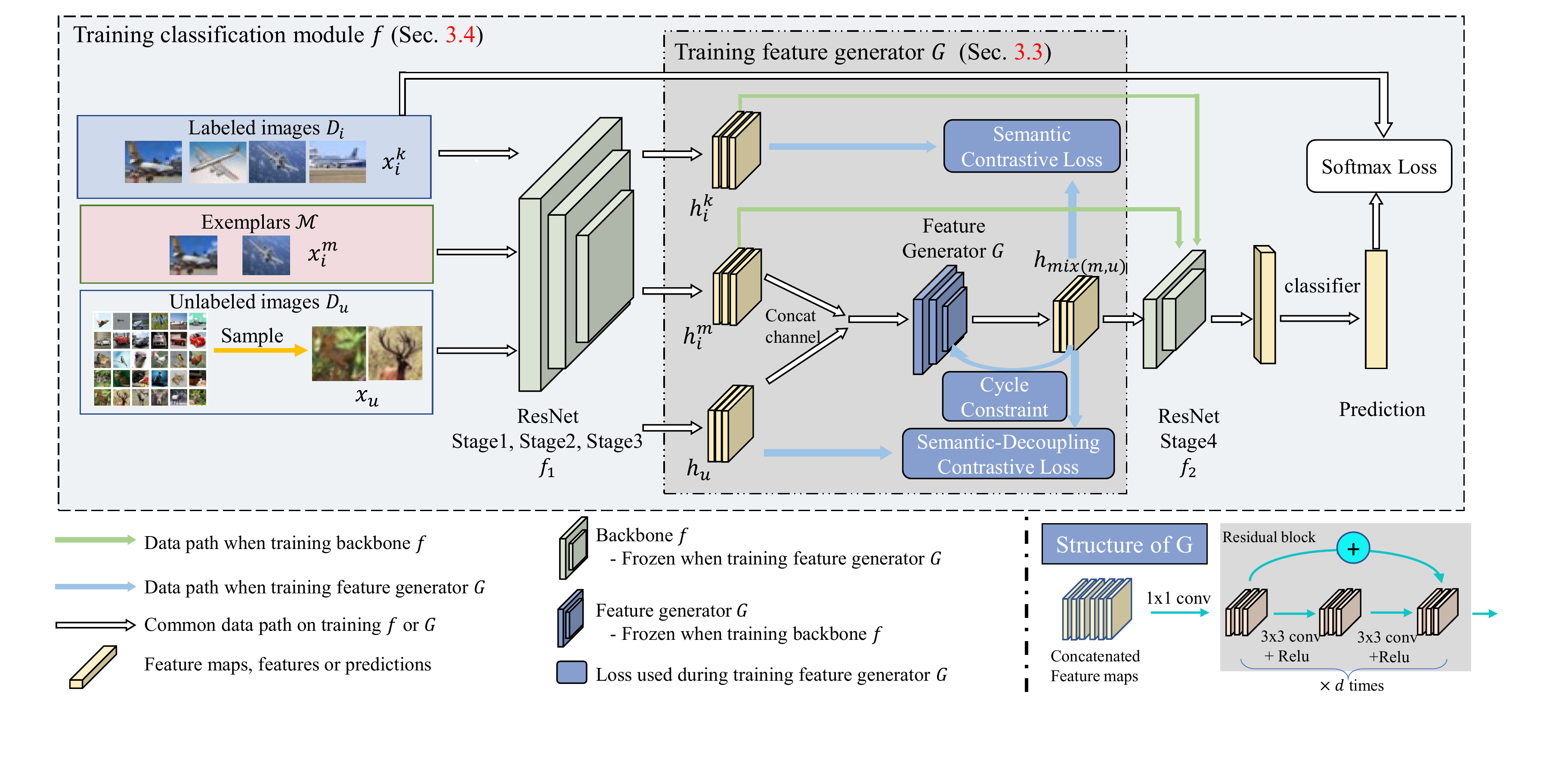}
    \caption{The training framework of the proposed method. At the end of each task, we create a set of feature generators $\{G\}$ and train the generators with exemplars, new images, and unlabeled images. In this process, semantic contrastive loss, semantic-decoupling contrastive loss, and cycle constraint are used to supervise $G$ for desired generated feature maps.
    We use lines in different colors to indicate data flows inside the model on different training proposes. The detailed structure of the feature generator $G$ is shown at the right-down corner. We adopt $d$ residual blocks for one $G_c$ and a $1\times1$ convolution layer right after residual blocks for channel fusion. Best viewed in color.}
  \label{fig:method}
\end{figure*}

\noindent\textbf{Structure-driven methods.} Popular structure-driven methods\cite{der,itaml,random_path,ccgn,CEC,bit,bic} modify the network structure or expand the network for training new tasks.
BiC\cite{bic} applies a linear layer to correct the classification bias via a small subset of exemplars.
DER\cite{der} expands the feature extracting backbone at each training step and tries to reduce parameters via pruning by a learnable channel-level mask.
iTAML\cite{itaml} starts with multiple task-specific models and utilizes meta-learning to better ensemble different models.
CCGN\cite{ccgn} plugs a gating structure for each convolution and fully-connecting layer to capture class-specific knowledge.
RPS-Net\cite{random_path} defines parallel modules at every layer and that forms a possible searching space that contains previous task-specific knowledge.
CEC\cite{CEC} designs a graph model for combining classifiers from different tasks.

Compared with structure-driven methods which often expand the network structure and need further finetune over these extra parameters, our introduced feature generator does not need to be further finetuned during training stages and brings no extra inference cost.

\section{Learning to Imagine against Forgetting}
\label{sec:method}

\subsection{Overview}
\noindent\textbf{Problem statement.}
For class incremental learning for a deep model $\Phi(f(\cdot))$ consisting of a feature extraction backbone $f$ and a classifier $\Phi$, we aim to continuously learn the deep model from a data stream of $N_T$ tasks denoted as $\{\mathcal{T}_i\}_{i=1}^{N_T}$. 
Since different tasks are totally disjoint, the classifier will grow to adapt to more classes 
when the number of tasks increases. 
Particularly, for the $i$-th task $\mathcal{T}_i$, the classification model is learned to classify a certain set of classes $\mathcal{Y}_i$ with a corresponding image dataset $\mathcal{D}_i = \{{x_i^k}, y_i^k\}_{k=1}^{N_i}$ as training data, where ${x_i^k}$ is the $k$-th image at task $\mathcal{T}_i$ with corresponding label $y_i^k \in \mathcal{Y}_i$ and $N_i$ is the size of the dataset. 
Once the learning task $\mathcal{T}_i$ finishes, the current dataset $\mathcal{D}_i$ will be discarded.
To prevent the model from forgetting knowledge about $\mathcal{Y}_i$ when learning new tasks, rehearsal based methods\cite{bic,icarl,podnet,DDE} try to keep a small portion of $\mathcal{D}_i$ denoted as $\mathcal{M}_i$ in advance, called `exemplars'.
After learning all the tasks, the model is supposed to perform well on all seen categories $\mathcal{Y}=\cup_i\mathcal{Y}_i$. 

\noindent\textbf{Challenge and idea.}
However, the limited exemplars are insufficient to remind the model of old knowledge and therefore the catastrophic forgetting remains.
To alleviate this problem, we develop a plug-and-play learnable feature generator $G$ to generate diverse counterparts of the exemplars by exploiting abundant semantically-irrelevant information within unlabeled data denoted as $\mathcal{D}_u$. And then, these diverse generated samples are used to train the classification model in order to keep the model's ability of classifying all seen classes.  
An overview is shown in \cref{fig:method}. 

\subsection{Learning Framework}
We form a two-step training framework: one step to train feature generator and another step to train network with the help of the generator.
Specifically, instead of constructing a unified generator for all seen classes, we develop a light-weight feature generator $G_c$ for each class $c$ to better capture the class-specific information.
When the $i$-th task ends, we use original dataset $\mathcal{D}_i$, exemplars $\mathcal{M}_i$ as well as unlabeled dataset $\mathcal{D}_u$ to learn class-specific feature generators before dropping $\mathcal{D}_i$. The generators are trained to generate diverse counterparts for $\mathcal{M}_i$ based on $\mathcal{M}_i$ and $\mathcal{D}_u$.
Here, we use the original data $\mathcal{D}_i$ to enforce that the generated samples should be semantically consistent and diverse. Besides, semantic-decoupling contrastive learning is applied to facilitate the exploration of the unlabeled dataset (\cref{sec:featurefusion}). 

When learning new tasks, the generators are used to help the classification model to overcome forgetting.
Since the feature map extracted from deeper layers is more likely to contain task-specific information\cite{visualizeCNN}, the deeper layer is easier to forget previous knowledge. 
Therefore, we split the feature extraction network $f$ into two parts: $f_1$ and $f_2$, \ie, $f(\cdot) = f_2(f_1(\cdot))$ and put our feature generators between $f_1$ and $f_2$.
The generators are frozen and generate diverse counterparts of exemplars by mixing the feature maps of exemplars and that of unlabeled data extracted by the bottom layer $f_1$.
These diverse generated samples will be used to remind the classification model of the old knowledge by guiding the model to correctly classify them  (\cref{sec:lwf}).

We first detail the learning of our feature generators in \cref{sec:featurefusion}, and then apply them to assist the classification model to overcome forgetting in \cref{sec:lwf}.

\subsection{Adaptive Feature Generator}
\label{sec:featurefusion}

Given a memory $\mathcal{M}_i$ and an original dataset $\mathcal{D}_i$, our goal is to learn a set of class-specific adaptive feature generators $\mathcal{G}_i$ that is capable of generating diverse samples which are semantically consistent in $\mathcal{M}_i$ by exploiting valuable unlabeled data. 
Formally, given an exemplar $x^m_{i} \in \mathcal{M}_i $ belonging to class $c$ (\ie, $y^m_i = c$) and an unlabeled sample $x_u$, we first extract their feature maps $h^m_i$ and $h_u$ through the bottom layers $f_1$ as:
\begin{equation}
    h^m_i = f_1(x^m_i),
\end{equation}
\begin{equation}
    h_u = f_1(x_u).
\end{equation}
And then, the class-specific feature generator $G_c$ is supposed to generate a feature map $h_{mix(m,u)}$ by adaptively mixing feature map from exemplars and the feature map from unlabeled data:
\begin{equation}\label{eq:generate}
    h_{mix(m,u)} = G_c(h^m_{i}, h_{u}).
\end{equation} 
Ideally, $G_c$ should have two properties: (i) preserving the semantic information from $h^m_{i}$ and (ii) augmenting $h^m_{i}$ with the semantically-irrelevant information from $h_u$.
Therefore, for (i), we apply semantic contrastive learning between $h_{mix(m,u)}$ and its intra-class counterparts. For (ii), we perform semantic-decoupling contrastive learning between $h_{mix(m,u)}$ and the unlabeled sample.

\noindent\textbf{Semantic contrastive learning.} To extract the semantic information from the mixed feature map $h_{mix(m,u)}$, we apply a global average pooling operation $GAP(\cdot)$ on $h_{mix(m,u)}$:
\begin{equation}\label{eq:gap}
    {v_{mix(m,u)}} = GAP(h_{mix(m,u)}).
\end{equation}
After encoding semantic information into vector $v_{mix(m,u)}$, we enforce the generated samples to have the same semantic information with $x^m_i$ so as to lead the feature generator $G_c$ to capture and preserve semantic information within the input feature map ${h^m_i}$. Here, instead of directly extracting semantic information from $h^m_i$ for supervision, we exploit semantic information from another sample $x^{k}_i$ from $\mathcal{D}_i$ that belongs to the same class with $x^m_i$ (\ie, $y^{k}_i = y^m_i = c$). 
Note that $x^k_i$ is sampled from $\mathcal{D}_i$, which forms the original data distribution with abundant intra-class variance, while $x^m_i$ is sampled from memory $\mathcal{M}_i$ in which the data distribution is often biased due to the small quantity of exemplars. Therefore, using $x_i^k$ to guide the generator will further close the gap between the distribution of generated samples and the original data distribution.
Consequently, the \textbf{S}emantic \textbf{C}ontrastive learning loss can be formulated as:
\begin{equation}\label{eq:sc}
    \mathcal{L}_{SC} = \| v^{k}_{i} - v_{mix(m,u)} \|_2,
\end{equation}
where $\|\cdot\|_2$ denotes the  ${L}_2$-norm and $v^{k}_i$ is the semantic information extracted from $x^{k}_i$ similar to \cref{eq:gap}.

\noindent\textbf{Semantic-Decoupling contrastive learning.}
To guide feature generator $G$ to `imagine' diverse samples referring to unlabeled data, we need to mine semantically irreverent information from unlabeled data.
To facilitate the exploration of the unlabeled dataset, we decouple the semantic information and mine the semantically irreverent information using gram matrix\cite{gram}.
The gram matrix is adopted  on the level of feature map to encode such semantically-irrelevant information following prior works\cite{gram,gram2,gram3,gram4,gram5}.
Formally, given a feature map $h_u$ of an unlabeled sample, its gram matrix could be calculated as:
\begin{equation} \label{eq:w}
    W_{u} = Gram({h_{u}}), 
\end{equation}
where $Gram(\cdot)$ denotes the inner product of the flattened vectors between pairwise channels.  
The value of position $(i,j)$ in ${W_{u}}$ is calculated as:
\begin{equation}
    W_u^{(i,j)} = Gram({h_u}) = {r_u^i}^Tr_u^j,
\end{equation}
where ${r_u^i} = flatten({h_u(i, :, :)})$ represents the flattened vector of $i$-th channel of feature map ${h_u}$ and $W_u$ encodes the relationships between channels.
By modeling the channel-wise relationships, we could obtain abundant semantically-irrelevant information such as the textures~\cite{gram,gram2,gram3,gram4,gram5}.
Similarly, we could compute the gram matrix of the generated samples $h_{mix(m,u)}$. A \textbf{S}emantic-\textbf{D}ecoupling \textbf{C}ontrastive loss is imposed to guide $G_c$ to learn the semantically-irrelevant information:
\begin{equation}\label{eq:dsc}
    \mathcal{L}_{SDC} = \|W_{u} - W_{mix(m, u)}\|_F,
\end{equation}
where $\|\cdot\|_F$ denotes the Frobenius norm.

\noindent\textbf{Cycle constraint.}
The semantic inputs of $G_c$ only come from the exemplars in $\mathcal{M}_i$ and the limited quantity of $\mathcal{M}_i$ may lead to the overfitting of the feature generator.
To facilitate the training process, we introduce a cycle constraint to further utilize the generated samples.
Specifically,  after we obtain the generated feature map ${h_{mix(m, u)}}$, we feed ${h_{mix(m, u)}}$ into $G_c$ to provide semantic information so that $G_c$ could learn to extract semantic information for not only exemplars but also generated samples, and therefore improve the generalization ability of generator $G_c$,:
\begin{equation}\label{eq:cycle_generate}
    h_{cyc(mix, m)} = G_c({h_{mix(m, u)}}, {h^m_i}).
\end{equation}

Here, we guide $G_c$ to extract the semantically-irrelevant information from exemplars $h^m_i$. Since we have guided  $G_c$ to extract semantic information from $h^m_i$ in \cref{eq:generate}, further encouraging $G_c$ to extract semantically-irrelevant information from $h^m_i$ in \cref{eq:cycle_generate} could impose a feature disentanglement constraint on $G_c$, which is beneficial for generating acquired feature maps.
We enforce the generator to extract semantic information from ${h_{mix(m, u)}}$ and decompose semantic information from ${h^{m}_i}$ in the same way as formulated in \cref{eq:sc} and \cref{eq:dsc}:
\begin{equation}
    \mathcal{L}_{SC}^{cyc} = \| v_{cyc(mix, m)} - v_{cyc(m, u)} \|_2,
\end{equation}
\begin{equation}
    \mathcal{L}_{SDC}^{cyc} = \|W_{cyc(mix, m)} - W^m_i\|_F,
\end{equation}
where $v_{cyc(mix, m)}, v_{cyc(m, u)}$ are obtained by \cref{eq:gap} and $W_{cyc(mix, m)}, W^k_i$are  based on \cref{eq:w}.
If the generator minimizes $\mathcal{L}_{SC}$ by overfitting on exemplars $h^m_i$, it will suffer from large loss due to the existence of $\mathcal{L}_{SC}^{cyc}$ and $\mathcal{L}_{SDC}^{cyc}$.

\noindent\textbf{Training objective of $G$.}
With all losses mentioned above, we have the following over all training loss
for module $G_c$:
\begin{equation}
    \mathcal{L}_G = \mathcal{L}_{ce} +  \mathcal{L}_{SC} + \lambda \mathcal{L}_{SDC} +
    \lambda_{cyc} (\mathcal{L}_{SC}^{cyc} + \lambda \mathcal{L}_{SDC}^{cyc}),
\end{equation}
where $\lambda$ and $\lambda_{cyc}$ are the trade-off parameters.  $\mathcal{L}_{ce}$ is the cross entropy loss used to ensure the discrimination of generated feature maps and is formulated as:
\begin{equation}
     \mathcal{L}_{ce} = - \log \Phi (f_2(h_{mix(m, u)}))^{(y^m_i)},
\end{equation}
where $\Phi (f_2(h_{mix(m, u)}))^{(y^m_i)}$ is the $y^m_i$-th element of $\Phi (f_2(h_{mix(m, u)}))$, representing the probability of $h_{mix(m, u)}$ belonging to class $y^m_i$.
Please note that $\mathcal{L}_G$ is the loss for single triplet samples $({x^m_{i}, {x_u}, x^{k}_{i}})$, and the total loss of a mini-batch is the average of $\mathcal{L}_G$ of all triplets within the batch. During the training process of $G_c$, we freeze network $f$ and $\Phi$.

\subsection{Anti-forgetting Learning}
\label{sec:lwf}
When learning a new task, we train the classifier $\Phi$ and backbone $f$ with the corresponding training data $\mathcal{D}_i$. 
For any samples from $\mathcal{D}_i$, we pass it through our backbone $f$ and the classifier $\Phi$, guiding the model correctly classify each sample in the new training set.
Formally, the classification loss can be formulated as:
\begin{equation}
    \mathcal{L}_{cls} = -\sum_{k=1}^{|\mathcal{D}_i|} \log \Phi(f({x^k_i}))^{(y_i^k)},
\end{equation}
where $\Phi(f({x_i^k}))^{(y_i^k)}$ is the $y_i^k$-th element of $\Phi(f({x_i^k}))$.

To overcome forgetting previous tasks, we not only use realistic data from  $\mathcal{D}_i$, and $\mathcal{M}=\cup_{i'=1} ^{i-1} \mathcal{M}_{i'}$ for training, but also use the generated feature maps. Since $\mathcal{M}$ denotes the union of exemplars from multiple previous tasks, we use $x^m$ \footnote{In the following, we do not care about the task label of exemplar, so we omit the task label of each exemplar in this section for simplicity.} denote the $m$-th instance in $\mathcal{M}$.

We utilize the feature generators to generate $n$ diverse counterparts for each exemplar based on $n$ unlabeled data in each training batch.
Specifically, we feed memorized exemplar ${x^m}$ (whose label is $y^m$) and $n$ unlabeled samples into $f_1$ and get outputs ${h^m} = f_1(h^m)$ and $\{h_u\}_{u=1}^n$.
And then, diverse counterparts of $x^m$ are obtained via feature generator $G_{y^m}$, denoted as $\{h_{mix(m,u)}\}_{u=1}^n$ where  $ h_{mix(m,u)}= G_{y^m}(h^m, h_u)$.
Without losing generality, we talk about the situation of $n=1$, \ie generating a counterpart for each exemplar $x^m$.
To train the network with diverse memory (the generated samples), we put ${h_{mix(m, u)}}$ into $f_2$ and classifier $\Phi$ to get prediction $\Phi(f_2({h_{mix(m, u)}}))$.
Finally, we compute a cross entropy loss for all exemplars and the generated sample $h_{mix(m, u)}$, whose ground truth label is consistent with $y_m$ as the feature generator $G_{y^m}$ had been trained to preserve the semantic information from $x^m$:
\begin{equation}
    \mathcal{L}_{cls}^{M} = -\sum_{m=1}^{|\mathcal{M}|} \log \Phi(f({x^m}))^{(y^m)},
\end{equation}
\begin{equation}
    \mathcal{L}_{cls}^{G} = -\sum_{m=1}^{|\mathcal{M}|} \sum_u \log  \Phi(f_2({h_{mix(m, u)}}))^{(y^m)},
\end{equation}
where $\Phi(f_2({h_{mix(m, u)}}))^{(y^m)}$ is the $y^m$-th element of $\Phi(f_2({h_{mix(m, u)}}))$.

\begin{table*}[]
\centering
\footnotesize
\begin{tabular}{c|cccccc|cccc}
    \toprule[2pt]
\multirow{3}{*}{Method} & \multicolumn{6}{c|}{CIFAR-100} & \multicolumn{4}{c}{ImageNet-Subset} \\ \cline{2-11} 
 & \multicolumn{2}{c|}{2 steps} & \multicolumn{2}{c|}{5 steps} & \multicolumn{2}{c|}{10 steps} & \multicolumn{2}{c|}{5 steps} & \multicolumn{2}{c}{10 steps} \\
 & Paras(M) & \multicolumn{1}{c|}{Avg Acc.} & Paras(M) & \multicolumn{1}{c|}{Avg Acc.} & Paras(M) & Avg Acc. & Paras(M) & \multicolumn{1}{c|}{Avg Acc.} & Paras(M) & Avg Acc. \\ \hline
Upper Bound & 0.46 & \multicolumn{1}{c|}{72.80} & 0.46 & \multicolumn{1}{c|}{72.80} & 0.46 & 72.80 & 11.2 & \multicolumn{1}{c|}{81.20} & 11.2 & 81.20 \\ \hline
\multicolumn{11}{l}{\textit{Parameter-growing methods} \ \ \textit{Memory size = $2000$}}  \\ \cdashline{1-11}
DER(w/o P)\cite{der} & \textgreater{}0.92 & \multicolumn{1}{c|}{70.18} & \textgreater{}1.61 & \multicolumn{1}{c|}{68.52} & \textgreater{}2.76 & 67.09 & - & \multicolumn{1}{c|}{-} & \textgreater{}67.2 & 78.20 \\
DER(P)\cite{der} & \textgreater{}0.32 & \multicolumn{1}{c|}{69.52} & \textgreater{}0.59 & \multicolumn{1}{c|}{67.60} & \textgreater{}0.61 & 66.36 & - & \multicolumn{1}{c|}{-} & \textgreater{}8.87 & 77.73 \\ \hline
\multicolumn{11}{l}{\textit{Parameter-static methods} \ \ \ \ \ \ \textit{Memory size = $2000$}} \\ \cdashline{1-11}
Icarl\dag\cite{icarl} & 0.46 & \multicolumn{1}{c|}{55.29} & 0.46 & \multicolumn{1}{c|}{56.29} & 0.46 & 52.42 & 11.2 & \multicolumn{1}{c|}{65.04} & 11.2 & 68.72 \\
DMC\dag\cite{dmc} & 0.46 & \multicolumn{1}{c|}{43.90} & 0.46 & \multicolumn{1}{c|}{38.20} & 0.46 & 23.80 & 11.2 & \multicolumn{1}{c|}{43.07} & 11.2 & 30.30 \\
GD\dag\cite{inthewild} & 0.46 & \multicolumn{1}{c|}{62.62} & 0.46 & \multicolumn{1}{c|}{56.39} & 0.46 & 51.30 & 11.2 & \multicolumn{1}{c|}{58.70} & 11.2 & 57.70 \\
BiC\dag\cite{bic} & 0.46 & \multicolumn{1}{c|}{48.43} & 0.46 & \multicolumn{1}{c|}{48.20} & 0.46 & 44.58 & 11.2 & \multicolumn{1}{c|}{70.07} & 11.2 & 64.96 \\
UCIR\dag\cite{ucir} & 0.46 & \multicolumn{1}{c|}{66.76} & 0.46 & \multicolumn{1}{c|}{59.66} & 0.46 & 55.77 & 11.2 & \multicolumn{1}{c|}{70.84} & 11.2 & 68.09 \\
TPCIL\cite{tpcil} & - & \multicolumn{1}{c|}{-} & - & \multicolumn{1}{c|}{65.34} & - & {63.58} & - & \multicolumn{1}{c|}{76.27} & - & 74.81 \\
Mnemonics\cite{Mnemonics} & - & \multicolumn{1}{c|}{-} & 0.46 & \multicolumn{1}{c|}{63.34} & 0.46 & 62.28 & 11.2 & \multicolumn{1}{c|}{72.58} & 11.2 & 71.37 \\
PODNet\cite{podnet} & 0.46 & \multicolumn{1}{c|}{\color{blue}67.69} & 0.46 & \multicolumn{1}{c|}{64.83} & 0.46 & 64.03 & 11.2 & \multicolumn{1}{c|}{75.54} & 11.2 & 74.58 \\
DDE\cite{DDE} & - & \multicolumn{1}{c|}{-} & - & \multicolumn{1}{c|}{\color{blue}65.42} & - & \color{blue}64.12 & - & \multicolumn{1}{c|}{\color{blue}76.71} & - & {\color{blue}75.41} \\
MixUp\ddag\cite{mixup} & 0.46 & \multicolumn{1}{c|}{62.30} & 0.46 & \multicolumn{1}{c|}{61.83} & 0.46 & 58.13 & 11.2 & \multicolumn{1}{c|}{69.82} & 11.2 & 68.55 \\
Ours & 0.46 & \multicolumn{1}{c|}{\textbf{\textcolor{red}{69.50}}} & 0.46 & \multicolumn{1}{c|}{\textbf{\textcolor{red}{68.01}}} & 0.46 & \textbf{\textcolor{red}{66.47}} & 11.2 & \multicolumn{1}{c|}{\textbf{\textcolor{red}{77.20}}} & 11.2 & \textbf{\textcolor{red}{76.76}}\\ \hline
\multicolumn{11}{l}{\textit{Parameter-static methods} \ \ \ \ \ \  \textit{Memory size = $1000$}} \\ \cdashline{1-11}
UCIR\cite{ucir} & - & \multicolumn{1}{c|}{-} & 0.46 & \multicolumn{1}{c|}{61.68} & 0.46 & 58.30 & 11.2 & \multicolumn{1}{c|}{68.13} & 11.2 & 64.04 \\
PODNet\cite{podnet} & - & \multicolumn{1}{c|}{-} & 0.46 & \multicolumn{1}{c|}{61.40} & 0.46 & 58.92 & 11.2 & \multicolumn{1}{c|}{\color{blue}74.50} & 11.2 & \color{blue} 70.40 \\
DDE\cite{DDE} & - & \multicolumn{1}{c|}{-} & - & \multicolumn{1}{c|}{\color{blue}64.41} & - & \color{blue}62.20 & - & \multicolumn{1}{c|}{71.20} & - & {69.05} \\
Ours & 0.46 & \multicolumn{1}{c|}{\textbf{\textcolor{red}{68.76}}} & 0.46 & \multicolumn{1}{c|}{\textbf{\textcolor{red}{67.08}}} & 0.46 & \textbf{\textcolor{red}{64.41}} & 11.2 & \multicolumn{1}{c|}{\textbf{\textcolor{red}{75.73}}} & 11.2 & \textbf{\textcolor{red}{74.94}}\\ \toprule[2pt]
\end{tabular}
\caption{Compare to state-of-the-art methods on CIFAR-100 and ImageNet-Subset. We conduct experiments on 3 different settings on CIFAR-100: 2, 5, and 10 steps, and we conduct 5 and 10 steps experiments on ImageNet-Subset. Paras(m) indicates the parameters used in inference after learning (counted in millions). We mark the best results in \textbf{\textcolor{red}{red}} and second-best results in \textcolor{blue}{blue} among all \emph{parameter-static} methods at the same memory budget. \dag denotes the results produced using the public authorized codes. \ddag denotes the experimental results produced by replacing our feature generator with the widely used MixUp\cite{mixup} data augmentation. Best viewed in color.}
\label{table:all}
\end{table*}

\noindent\textbf{Training objective for classification model.} Objective function for training model  $\Phi(f(.))$ are summarized as:
\begin{equation}
    \mathcal{L} = \mathcal{L}_{cls} + \alpha_1 (\mathcal{L}_{cls}^{M} + \mathcal{L}_{cls}^{G} )+ \alpha_2 \mathcal{L}_{dist},
\end{equation}
    where $\alpha_1$ and $\alpha_2$ are the trade-off parameters.
    $\mathcal{L}_{dist}$ is the widely-used distillation loss in previous works~\cite{lwf,zhao2020maintaining,ucir,castro2018end}, which forces the current feature space  close to the old feature space to overcome forgetting and is formulated as:
    \begin{equation}
    \mathcal{L}_{dist} = \sum_{k=1}^{N_m}(1-\frac{{{{f}_{old}({x_m^k})}}\ ^T\ {{{{f}({x_m^k})}}}}{\|{{f}_{old}({x_m^k})}\|_2 \cdot \|{{{f}({x_m^k})}}\|_2}),
    \end{equation} 
    where $f_{old}$ is the old network backbone (of last task $\mathcal{T}_{n-1}$).

\section{Experiments}
In this section, extensive experiments are shown to validate the effectiveness of the 
proposed method. Experiment setup is list in \cref{sec:implemeted_details}. We compare our method with existing state-of-the-art incremental learning methods in \cref{sec:SOTA}. In \cref{sec:ablation_study}, we validate the indispensability of each objective function proposed for generator $G$. We further analyze each component of our method in \cref{sec:futher_analysis}.
More experiments including evaluations with different trade-off parameters are in Appendix.

\subsection{Experiment Setup}
\label{sec:implemeted_details}
\noindent\textbf{Datasets.} We conduct experiments on two widely used image classification
datasets: CIFAR-100\cite{cifar100} and ImageNet-Subset\cite{imagenet}. CIFAR-100 training set 
contains 100 classes and there are totally 50,000 images for training and 10,000 images for evaluation. ImageNet is a large-scale dataset containing 1,000 classes, 1.2 million images. For simplicity, we follow prior works\cite{ucir,podnet,tpcil, Mnemonics, DDE, der} and use ImageNet-Subset which contains 100 classes.

\noindent\textbf{Auxiliary Datasets.} As stated in \cref{sec:method}, our method is capable of adaptively generating diverse exemplars referring to unlabeled data.
We use ImageNet with different modifications following previous works \cite{dmc,inthewild} as our unlabeled dataset due to its diversity.
For CIFAR-100 dataset, to align the input resolution, we use $32\times32$ down-sampled ImageNet\cite{inthewild} as the auxiliary unlabeled dataset. 
For ImageNet-Subset which contains 100 classes, we use the rest 900 classes from original ImageNet as the auxiliary unlabeled dataset. 
We also evaluate our method with different auxiliary unlabeled data in \cref{sec:futher_analysis}.

\noindent\textbf{Testing Protocols.} We follow a popular testing protocol in class-incremental
learning\cite{podnet,ucir,icarl, der, DDE, tpcil, Mnemonics}. Experiments are started by training the model on 
half the classes, that is, 50 classes for CIFAR-100 and ImageNet-Subset.
The rest classes are added incrementally in steps. We split rest classes into 2, 5, 10
steps to validate our method. After each training task, the model will be evaluated by 
testing on all classes that had been seen until the current task and each 
top-1 accuracy in every task is averaged for a final score called \textit{average accuracy}~\cite{icarl,podnet,tpcil,der,DDE}.
Following prior works~\cite{icarl,podnet,der,ucir}, we limit our memory budget to 2,000 exemplars for 100-classes datasets (including CIFAR-100 and ImageNet-Subset).

\noindent\textbf{Implementation Details.} For CIFAR-100, we adopt a modified 32-layers ResNet as in previous works \cite{icarl, podnet, der, DDE}, which have fewer channels and shallower stages compared to the official ResNet-32\cite{resnet, der}. 
For ImageNet-Subset, we use the original RseNet-18 as our backbone following prior works\cite{podnet,der,tpcil, DDE, Mnemonics, bic}.
As for the memory saving strategy, we follow a popular herding selection strategy proposed in \cite{icarl}.
We use SGD optimizer with an initial learning rate of 0.1. The feature generator $G$ is plugged between stage3 and stage4 of the ResNet.

\subsection{Comparison to the State-of-the-Art}
\label{sec:SOTA}

We conduct experiments on CIFAR-100 and ImageNet-Subset and compare our method to state-of-the-art methods\cite{icarl,ucir,podnet,dmc,inthewild,bic,ucir,tpcil,Mnemonics,DDE}.
For CIFAR-100, we incrementally learn the classes in 2,5 and 10 steps to better valid our method following previous works~\cite{der,podnet, DDE, tpcil, Mnemonics}.
For ImageNet-Subset, we divide the classes equally into 5, 10 incremental tasks to validate our method following previous works~\cite{der,podnet, DDE}.

Note that DER\cite{der} has increasing parameters for inference during incremental learning processes and we categorize it as parameter-growing method.
Differently, our method has fixed and fewer parameters for inference since our inference does not rely on the feature generator, so we categorize our method as parameter-static method. 
We also compare our method with other parameter-static methods, such as PODNet\cite{podnet}, DDE\cite{DDE} and TPCIL\cite{tpcil}. 
To show the upper bound for reference, we train a model without splitting the training set and evaluate it on the test set.
Moreover, we evaluate our method under a more limited memory budget to further show the effectiveness of our method.

\noindent\textbf{Results on CIFAR-100.} \ \ 
Our method outperforms other parameter-static methods by a clear margin under all the experimental settings as shown in \Cref{table:all}.
Under the setting of 10 steps incremental learning, our method achieves $66.47\%$ accuracy, which is about 3\% higher than advanced \emph{parameter-static} methods PODNet, UCIR, BiC, DDE and Icarl.
Surprisingly, our method also outperforms advanced \emph{parameter-growing} method DER(P) by 0.11\% accuracy with fewer parameters (0.46M vs. more than 0.61M). 
It is mainly because our method could effectively utilize the abundant semantically-irrelevant information within unlabeled data to adaptively diversify exemplar, helping the model overcome forgetting of old knowledge.

When we cut the memory budget down to half, our method still performs best among parameter-static methods and even outperforms their counterparts which have 2000 exemplars. For example, we achieve 67.08\% accuracy under the 5-step setting with 1000 exemplars, which is 1.66\% higher than that of the DDE with 2000 exemplars. 

To further demonstrate the importance of the learnable feature generator, we replace generator $G$ with widely used data augmentation MixUp\cite{mixup} and the experimental results demonstrate the superiority of our method. 
The reason is that MixUp will inevitably destroy the semantic information from exemplars since it coarsely mixes two images at pixel level. Differently, our proposed method is learnable, making it more effective at exploiting semantically-irrelevant information from unlabeled data while keeping the generated samples semantically consistent with exemplars.

\noindent\textbf{Results on ImageNet-Subset.} \ \ 
Our method outperforms advanced parameter-static methods and is comparable to advanced parameter-growing method DER as shown in \Cref{table:all}.
Under the 5-step setting, our method achieves 77.2\% accuracy, which is 1.66\% higher than that of PODNet.
When there are more learning steps or fewer exemplars, the performance gap becomes larger, showing that our method is more effective in overcoming forgetting. 
For example, our method achieves 76.76\% accuracy under the 10-step setting, which is 2.18\% higher than PODNet, and such the performance gap becomes 4.54\% when there are only 1000 exemplars.
It is mainly because previous works are limited by the small number of exemplars while our method could effectively diversify the exemplars.
Besides, our method outperforms MixUp by a large margin as in CIFAR100.

The experiments under 1000 exemplars suggest that our method is superior to others since we could achieve comparable performance to others with fewer exemplars, which helps to address the limitations such as privacy concerns.

\subsection{Ablation Study}
\label{sec:ablation_study}
\begin{table}[]
    \centering
    \footnotesize
    \begin{tabular}{c|c}
    \toprule[1.3pt]
    Method & Avg Acc. \\ \hline
    Baseline & 63.98 \\
     + $\mathcal{L}_{SC}$ & 64.41 \\
     + $\mathcal{L}_{SC}$ + $\mathcal{L}_{SDC}$ & 65.53 \\ 
     + $\mathcal{L}_{SC}$ + $\mathcal{L}_{SDC}$ +  $\mathcal{L}_{SC}^{cyc}$ + $\mathcal{L}_{SDC}^{cyc} $ \ & 66.47  \\ 
    \toprule[1.3pt]
    \end{tabular}
    \caption{Effectiveness of each objective function during training $G$. `Baseline' denotes training model without $G$ while other methods use $G$ for feature generator.  
    Experiments are conducted on CIFAR-100 under the 10 step incremental setting.
    }
    \label{table:ablation}
    \end{table}
In this section, we discuss the effectiveness of each objective function for $G$. 
We perform ablation studies on CIFAR-100 using modified ResNet-32 as backbone under 10 steps incremental learning to analyze the effect of each component in $G$.
Our baseline is to train the deep model with the task-specific training dataset $\mathcal{D}_i$ and the limited exemplar memory $\mathcal{M}$ in each task.

\noindent\textbf{The effectiveness of the semantic contrastive learning.}
From \Cref{table:ablation}, we can observe that with the help of $\mathcal{L}_{SC}$, 
our method outperforms baseline by 0.5\%. It is mainly because the $\mathcal{L}_{SC}$ encourages generated samples to be semantically consistent with $\mathcal{L}_{SC}$ and improve the performance. 

\noindent\textbf{The effectiveness of the semantic-decoupling contrastive learning.}
As shown in \Cref{table:ablation}, when further combining $\mathcal{L}_{SC}$ with $\mathcal{L}_{SDC}$ to train the model, the model outperforms the baseline model by 1.55\% accuracy. It is mainly because  the feature generator $G$ is explicitly guided to decompose semantic information from unlabeled data by $\mathcal{L}_{SDC}$, leading to more diverse generated samples.

\noindent\textbf{The effectiveness of the cycle constraint.}
Upon the $\mathcal{L}_{SC}$ and $\mathcal{L}_{SDC}$, applying cycle constraint to our model could further boost the performance from 65.53\% to 66.47\% at accuracy as shown in \Cref{table:ablation}. It is mainly because the cycle constraint could effectively prevent the overfitting of the generator $G$, making the generated samples more diverse.

\subsection{Further Analysis}
\label{sec:futher_analysis}

\begin{table}[]
    \centering
    \footnotesize
    \begin{tabular}{c|c}
    \toprule[1.3pt]
    Unlabeled dataset & Avg Acc. \\ \hline
    ImageNet1k & 66.47 \\
    ImageNet-ec &  66.11\\
    ImageNet-500 & 66.07 \\
    ImageNet-300 & 65.55 \\ 
    ImageNet-100 & 64.23 \\ 
    \toprule[1.3pt]
    \end{tabular}
    \caption{The impact of different unlabeled datasets when incrementally learning CIFAR-100. 
    `ImageNet-ec' indicates the classes that appear in CIFAR-100 are excluded from ImageNet1k.
    We form several subsets by sampling different number of classes from `ImageNet-ec', denoted as `ImageNet-\#classes'.
    }
    \label{table:unlabeled_dataset}
    \end{table}

\noindent\textbf{Investigation on unlabeled data.}
In the last section, we use a resized version of ImageNet1k as the auxiliary unlabeled dataset.
To show that the performance gain is not mainly from the overlapping classes between the ImageNet1k and CIFAR-100, we exclude the classes in CIFAR-100 from ImageNet1k and denote the resulting dataset as ImageNet-ec. The experimental result about using ImageNet-ec as unlabeled data is shown in \Cref{table:unlabeled_dataset}.
We find out that our method does not rely on the overlapping classes between unlabeled dataset and labeled dataset because the results denoted as `ImageNet-ec' are slightly lower than  `ImageNet-full' in \Cref{table:unlabeled_dataset}.

To further study the impact of unlabeled datasets with different scales, we evaluate our method using different subsets of ImageNet-ec.
Specifically, we select the first 100, 300, 500 classes of ImageNet-ec as the unlabeled datasets denoted as `ImageNet-\#classes'.
From the experimental results in \Cref{table:unlabeled_dataset}, we can observe that the performance drops 2.2\% when ImageNet-100 is used as unlabeled data, indicating that the abundant semantically-irrelevant information in the unlabeled dataset is critical to our method.
Besides, our performance becomes stable when the unlabeled dataset declines from  `ImageNet-ec' to `ImageNet-500'. It is mainly because our proposed feature generator is efficient at generating diverse samples for CIFAR-100 with abundant information in `ImageNet-500'.    

\begin{figure}
        \centering
        \begin{subfigure}{0.23\textwidth}
        \centering
            \includegraphics[width=\linewidth]{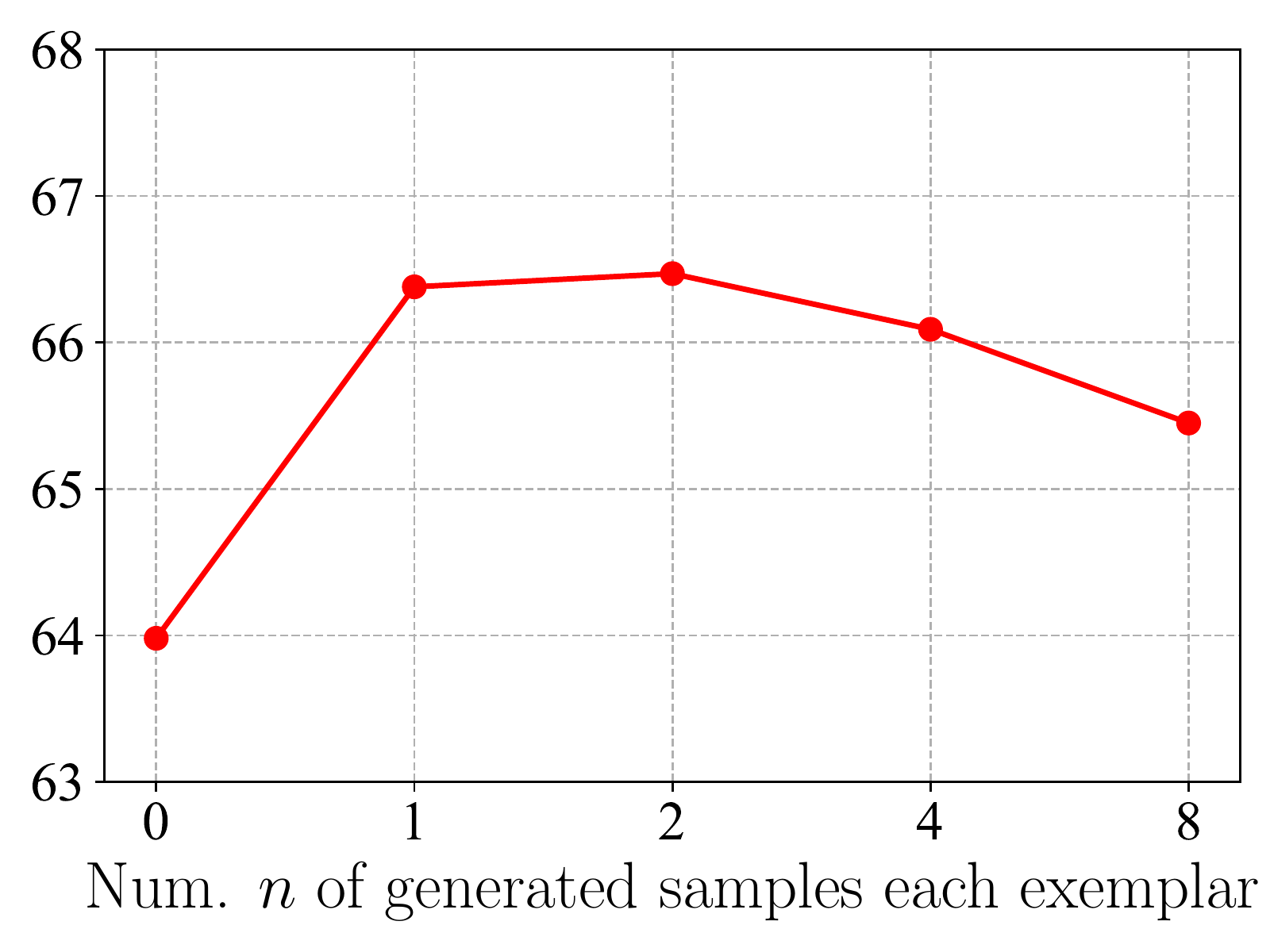}
            \caption{Experiment on the impact of the number of generated samples.}
            \label{fig:num_samples}
        \end{subfigure}%
        \ \ \
        \begin{subfigure}{0.23\textwidth}
        \centering
            \includegraphics[width=\linewidth]{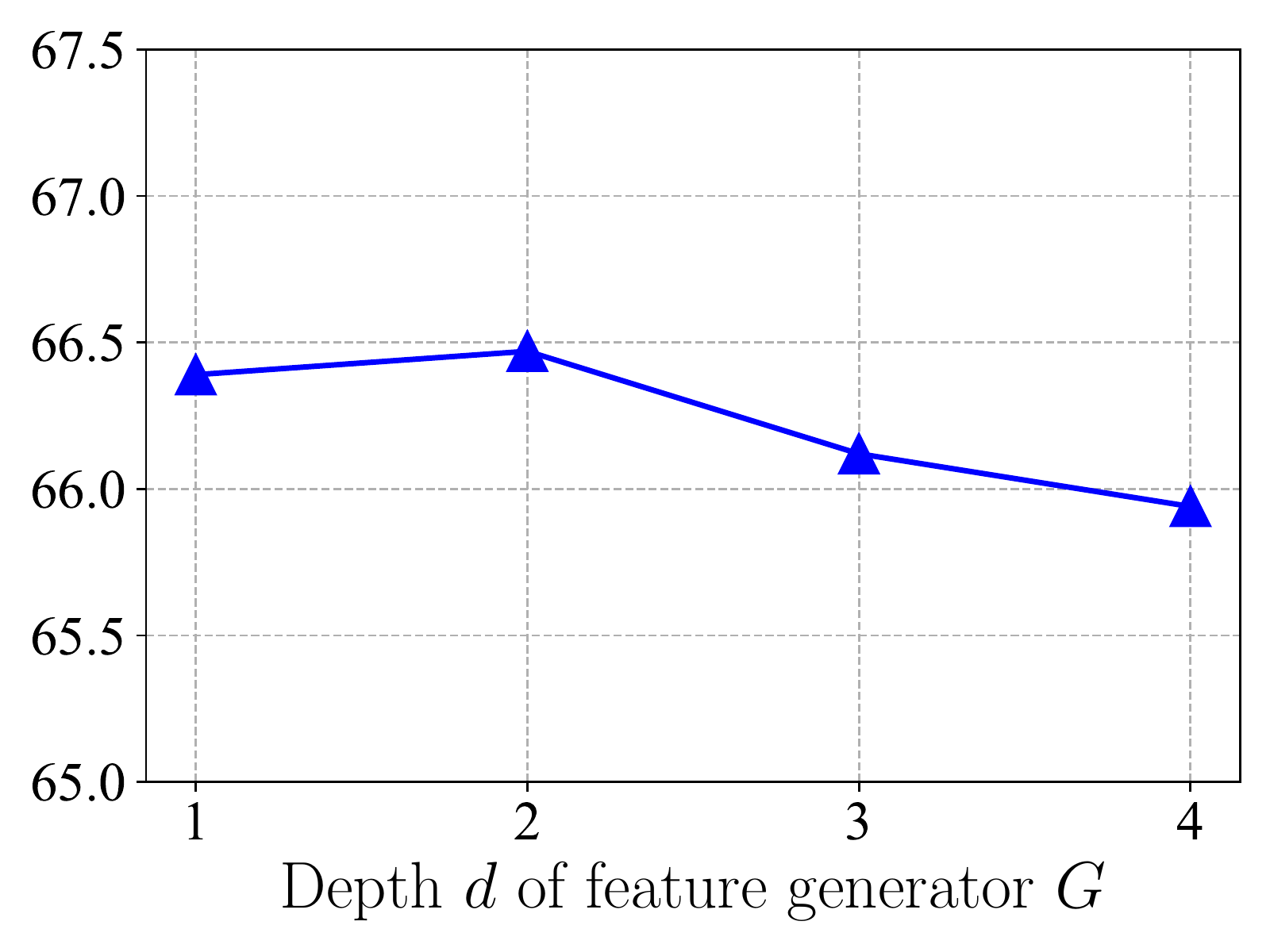}
            \caption{Experiment on the depth of feature generator $G$.}
            \label{fig:depth_G}
        \end{subfigure}
        \caption{We conduct experiments on the impact of the number of fused samples (a) and the impact of depth of feature generator $G$ (b). All experiments are conducted on CIFAR-100 and use ImageNet1k-32x32 as the unlabeled dataset.}
        \label{fig:anaysis}
        \end{figure}

\noindent\textbf{Number of generated samples.}
To investigate the impact of the number of generated counterparts of each exemplar in a training batch, denoted as $n$, we evaluate our method on CIFAR-100 with $n=0$ (baseline), $n=1$, $n=2$, $n=4$, and $n=8$. 
Results from \cref{fig:num_samples} show that when we adopt generated samples, even only generating one sample per exemplar in each training batch, the performance increases about 3\%, which indicates the effectiveness of our method.
Our method is not sensitive to $n \in [1, 4]$ and the optimal $n$ is $2$. 
When $n$ is greater than $4$, the performance drops significantly. 
This shows that our method could effectively generate diverse counterparts for exemplars by exploiting  unlabeled data. Meanwhile, when generating too many samples, the model will spend most of its capacity on previous knowledge, hindering its ability to learn new tasks.

\noindent\textbf{Structure of $G$.} To show the impact of the structure of $G$, we construct our feature generator $G$ with a different number of Residual blocks.
We set the number of Residual blocks $d$ to 1, 2, and 4. 
The experimental results shown in \cref{fig:depth_G} indicate that our feature generator $G$ performs well even with a single residual block and performs better at $d=2$ and our method is not sensitive to $d \in [1, 4]$.
When the feature generator $G$ gets deeper, the performance drops slightly. It is mainly because the $G_c$ is trained with limited exemplars (the input of $G_c$ are exemplars and the generated samples produced by itself), and a deeper $G_c$ will easily overfit these inputs and degenerate its performance.

\section{Conclusion}
In this paper, we argue that the small number of exemplars hinders overcoming forgetting. 
To this end, we proposed a learnable feature generator to generate diverse counterparts for the exemplars by adaptively mixing semantic-irreverent information from unlabeled data with semantic information from exemplars.
The generator is frozen after its training and generates diverse samples to remind the model of the old task. Moreover, the generator is not needed during inference, making our method efficient. Extensive experiments demonstrate that our method outperforms state-of-the-art methods in two widely used datasets.  
\section{Acknowledgment}
This work was supported partially by the NSFC (U21A20471, U1911401, U1811461), Guangdong NSF Project (No. 2020B1515120085, 2018B030312002), Guangzhou Research Project (201902010037), and the Key-Area Research and Development Program of Guangzhou (202007030004).
{\small
\bibliographystyle{ieee_fullname}
\bibliography{egbib}
}
\end{document}